\documentclass[runningheads]{llncs}
\usepackage{graphicx}

\usepackage{tikz}
\usepackage{comment} 
\usepackage{amsmath,amssymb} 
\usepackage{multirow}
\usepackage{ctable}
\usepackage{color,colortbl}

\definecolor{Gray}{gray}{0.9}
\newcommand{\tabincell}[2]{\begin{tabular}{@{}#1@{}}#2\end{tabular}}

\begin{document}
\pagestyle{headings}
\mainmatter
\def\ECCVSubNumber{100}  

\title{HMOR: Hierarchical Multi-Person Ordinal Relations for Monocular Multi-Person 3D Pose Estimation} 

\titlerunning{Hierarchical Multi-Person Ordinal Relations for 3D Pose Estimation}
%
\author{Jiefeng Li$^{\star}$\inst{1} \and
Can Wang\thanks{Denotes equal contribution.}\inst{2} \and
Wentao Liu\inst{2} \and
Chen Qian\inst{2} \and
Cewu Lu\inst{1}\thanks{Cewu Lu is the corresponding author. He is the member of Qing Yuan Research Institute and MoE Key Lab of Artificial Intelligence, AI Institute, Shanghai Jiao Tong University, China.}}
\authorrunning{Li et al.}
%
\institute{Shanghai Jiao Tong University, Shanghai, China \\
\email{\{ljf\_likit, lucewu\}@sjtu.edu.cn} \and
SenseTime Research, Beijing, China \\
\email{\{wangcan, liuwentao, qianchen\}@sensetime.com}
}
\maketitle

\begin{abstract}
  Remarkable progress has been made in 3D human pose estimation from a monocular RGB camera. However, only a few studies explored 3D multi-person cases. In this paper, we attempt to address the lack of a global perspective of the top-down approaches by introducing a novel form of supervision - \textit{Hierarchical Multi-person Ordinal Relations (HMOR)}. The HMOR encodes interaction information as the ordinal relations of depths and angles hierarchically, which captures the \textit{body-part} and \textit{joint} level semantic and maintains global consistency at the same time. In our approach, an integrated top-down model is designed to leverage these ordinal relations in the learning process. The integrated model estimates human bounding boxes, human depths, and root-relative 3D poses simultaneously, with a coarse-to-fine architecture to improve the accuracy of depth estimation. The proposed method significantly outperforms state-of-the-art methods on publicly available multi-person 3D pose datasets. In addition to superior performance, our method costs lower computation complexity and fewer model parameters.
\end{abstract}

\section{Introduction}

Estimating 3D human poses from a monocular RGB camera is fundamental and challenging. It has found applications in robotics~\cite{robotics2,robotics1}, activity recognition~\cite{activity0,activity4}, human-object interaction detection~\cite{fang2018eccv,qi2018learning,li2020detailed,li2020pastanet}, and content creation for graphics~\cite{avatar0,avatar1}. With deep neural networks~\cite{vgg,resnet,hourglass,pang2019deep} and large scale publicly available datasets~\cite{humaneva,h36m,mpii,mscoco,panoptic,3dhp,singleshot,3dpw}, significant improvement has been achieved in the field of 3D pose estimation. Most of the works~\cite{coarse,martinez2017simple,compositional,zhou2017towards,fang2018,integral,twostage4,chen2019camera} focus on estimating the single-person pose. Recently, some methods~\cite{lcr,singleshot,nips2018,zanfir2018monocular,lcrpp,pirinen2019domes,moon} start to deal with multi-person cases. However, recovering absolute 3D poses in the camera-centered coordinate system is quite a challenge. Since multi-person activities take place in cluttered scenes, inherent depth ambiguity and occlusions make it still difficult to estimate the absolute position of multiple instances.

Recently, top-down approaches~\cite{lcr,lcrpp,moon} achieve noticeable improvements in estimating multi-person 3D poses. These approaches first perform human detection and estimate the 3D pose of each person by a single-person pose estimator. However, the pose estimator is applied to each bounding box separately, which raises the doubt that the top-down models are not able to understand multi-person relationships and handle complex scenes. Without a broad view of the input scenario, it is challenging to get rid of inherent depth ambiguity and occlusion problems. In this paper, the relationship among multiple persons is fully considered to address this limitation of top-down approaches.


We propose a novel form of supervision for 3D pose estimation - \textit{Hierarchical Multi-person Ordinal Relations (HMOR)}. HMOR explicitly encodes the interaction information as ordinal relations, supervising the networks to output 3D poses in the correct order. Different from previous works~\cite{pavlakos2018ordinal,wang2018drpose3d,ronchi2018ordinal} that only use relative depth information, HMOR considers both depths and angles relations and expresses the ordinal information hierarchically, i.e., \textit{instance} $\rightarrow$ \textit{part} $\rightarrow$ \textit{joint}, which makes up for the lack of a global perspective of the top-down approaches.

Further, we propose an integrated top-down model to learn this knowledge by encoding it into the learning process. The integrated model can be end-to-end trained with back-propagation and performs \textit{human detection}, \textit{pose estimation}, and \textit{human-depth estimation} simultaneously. Since metric depth from a single image is fundamentally ambiguous, estimating absolute 3D pose suffers from inaccurate human-depth estimation. To improve the accuracy, we take a coarse-to-fine approach to estimate human depth: i) initializes a global depth map, and ii) finetunes the human depths by estimating the correction residual.


We evaluate our method on two multi-person~\cite{singleshot,panoptic} and one single-person  3D pose datasets~\cite{h36m}. Our method significantly outperforms previous multi-person 3D pose estimation methods~\cite{lcr,singleshot,nips2018,xnect,kolotouros2019learning,moon} by \textbf{12.3} PCK$_{abs}$ improvement on the MuPoTS-3D~\cite{singleshot} dataset, and \textbf{20.5} mm improvement on CMU Panoptic~\cite{panoptic} dataset, with lower computation complexity and fewer model parameters. Compared to state-of-the-art single-person methods~\cite{fang2018,compositional,integral,zhou2019hemlets}, our method does not need ground-truth bounding-box in the inference phase and still achieves comparable performance. Additionally, our proposed method is compatible with 2D pose annotations, which allows the 2D-3D mixed training strategy.

The contributions of this paper can be summarized as follows:
\begin{itemize}
   \item[$\bullet$] We propose HMOR, a novel form of supervision, to explicitly leverage the relationship among {multiple persons} for pose estimation. HMOR divides human relations into three levels: \textit{instance}, \textit{part} and \textit{joint}. This hierarchical manner ensures both the global consistency and the fine-grained accuracy of the predicted results.
   \item[$\bullet$] An integrated end-to-end top-down model is proposed for multi-person 3D pose estimation from a monocular RGB input. We design a coarse-to-fine architecture to improve the accuracy of human-depth estimation. Our model jointly performs human detection, human-depth estimation, and 2D/3D pose estimation.
\end{itemize}

\section{Related Work}

\noindent\textbf{Multi-person 2D Pose Estimation.}\quad
Most of the multi-person 2D pose estimation methods can be divided into two categories: bottom-up and top-down approaches. Bottom-up approaches localize the body joints and group them into different persons. Traditional top-down approaches first detect human bounding boxes in the image and then estimate single-person 2D poses separately.

Representative works~\cite{openpose,associative,multiposenet,jin2019multi} of the bottom-up approaches are reviewed. Cao et al.~\cite{openpose} propose part affinity fields (PAFs) to model human bones. Complete skeletons are assembled by detected joints with PAFs. Newell et al.~\cite{associative} introduce a pixel-wise tag to assign joints to a specific person. Kocabas et al.~\cite{multiposenet} assign joints to detected persons by a pose residual network.

Top-down approaches~\cite{rmpe,maskrcnn,cpn,xiu2018pose,simplepose,li2019crowdpose,hrnet} achieve impressive accuracy in multi-person 2D pose estimation. Mask R-CNN~\cite{maskrcnn} is an end-to-end model to estimate multiple human poses but still process multiple persons separately. Fang et al.~\cite{rmpe} propose a two-stage framework (RMPE) to reduce the effect of the inaccurate human detector. Sun et al.~\cite{hrnet} propose the HRNet that maintains high-resolution representations through the whole process.

~\\
\noindent\textbf{Single-person 3D Pose Estimation.}\quad
There are two approaches to the problem of single-person 3D pose estimation from monocular RGB: single-stage and two-stage approaches.

Single-stage approaches~\cite{coarse,3dhp,compositional,kanazawa2018end,integral} directly locate 3D human joints from the input image. For example, Pavlakos et al.~\cite{coarse} propose a coarse-to-fine approach to estimate a 3D heatmap for pose estimation. Kanazawa et al.~\cite{kanazawa2018end} recover 3D pose and body mesh by minimizing the reprojection loss. Sun et al.~\cite{integral} operate an integral operation as soft-argmax to obtain 3D pose coordinates in a differentiable manner.

Two-stage approaches~\cite{twostage6,twostage2,twostage5,twostage1,twostage3,martinez2017simple,zhou2017towards,fang2018,twostage4} first estimate 2D pose or utilize the off-the-shelf accurate 2D pose estimator, and then lift them to the 3D space. Martinez et al.~\cite{martinez2017simple} propose a simple baseline to regress 3D pose from 2D coordinates directly. Moreno-Noguer~\cite{twostage3} obtains more precise pose estimation by the distance matrix representation. Yang et al.~\cite{twostage4} utilize a multi-source discriminator to generate anthropometrically valid poses.

~\\
\noindent\textbf{Multi-person 3D Pose Estimation.}\quad
A few works explore the problem of multi-person 3D pose estimation from a monocular RGB. Rogez et al.~\cite{lcr,lcrpp} propose LCR-Net and LCR-Net++. They locate human bounding boxes and classify those boxes into a set of K anchor-poses. A regression module is proposed to refine the anchor-pose to the final prediction. Instead of using a learning-based manner, they obtain the human depth by minimizing the distance between the projected 3D pose and the estimated 2D pose. Mehta et al.~\cite{singleshot} propose a bottom-up method. Their proposed occlusion-robust pose-map (ORPM) enables full body pose inference even under strong partial occlusions. Zanfir et al.~\cite{nips2018} propose MubyNet, a bottom-up model. MubyNet integrates a limb scoring model and formulates the person grouping problem as an integer program. Moon et al.~\cite{moon} propose a top-down two-stage model. They utilize the off-the-shelf human detection model and then perform single-person 3D pose estimation and root-joint localization. Those top-down approaches are not able to utilize multi-person relations since they estimate individual 3D pose separately. The bottom-up approaches are still suffering from limited accuracy. Our method combines the advantages of both approaches and boosts multi-person absolute 3D pose estimation by leveraging the multi-person relations in the integrated end-to-end top-down model.

~\\
\noindent\textbf{Ordinal Relations.}\quad
In the context of computer vision, several works learn ordinal apparent depth~\cite{zoran2015ordinal,chen2016ordinal} or reflectance~\cite{narihira2015ordinal,zhou2015ordinal} relationship as weak supervision. They motivated by the fact that ordinal relations are easier for humans to annotate. In the case of single-person 3D pose estimation, ~\cite{pavlakos2018ordinal,ronchi2018ordinal,sharma2019ordinal} use depth relations of body joints to generate 3D pose from 2D pose.

\section{Method}

We propose a novel representation, \textit{Hierarchical Multi-person Ordinal Relation (HMOR)}, to explicitly leverage ordinal relations among multiple persons and improve the performance of 3D pose estimation. Compared with previous works~\cite{pavlakos2018ordinal,wang2018drpose3d,ronchi2018ordinal} that use ordinal relation in 3D pose estimation, HMOR extends this idea in three dimensions: i) single-person to multi-persons, ii) \textit{joint} level to hierarchical \textit{instance-part-joint} levels, iii) depth relations to angle relations. Further, we develop an integrated model to aggregate HMOR into the end-to-end training process. In this section, we first describe the unified representation of the absolute multi-person 3D pose recovery under the top-down framework (\S\ref{sec:representation}). Then we detail the encoding and training schemes of the proposed HMOR (\S\ref{sec:hmor}). Finally, the integrated model with a coarse-to-fine depth estimation design is elaborated (\S\ref{sec:end2end}).


\subsection{Representation}\label{sec:representation}

Our task is to recover multiple absolute 3D human poses $\mathcal{P} = \{ \mathbf{P}^{abs}_{m} \}_{m=1}^N$ in the camera-centered coordinate system, where $N$ denotes the number of persons in the input RGB image. We assume that there are ${J}$ joints in a single 3D pose skeleton. The $m^{\text{th}}$ absolute 3D pose can be formulated as:
\begin{equation}
    \mathbf{P}^{abs}_{m} = \{\mathbf{k}_{m,j} : ({x}^{abs}_{{m},j}, {y}^{abs}_{{m},j}, {z}^{abs}_{{m},j})^{\mathsf{T}}\}_{j=1}^J ,
\end{equation}
where $\mathbf{k}_{m,j}$ is the $j^{\text{th}}$ joint position of the $m^{\text{th}}$ absolute pose.

Human bounding boxes $\{ \hat{\mathbf{B}}_{m} \}_{m=1}^N$, root-relative 3D poses $\{ \hat{\mathbf{P}}^{\mathit{rel}}_{m} \}_{m=1}^N$, and absolute depth of the root-joint $\{ \hat{{z}}^{abs}_{m, R} \}_{m=1}^N$ are needed to estimate the absolute 3D poses. We term root-joint's absolute depth as human depth, corresponding to the pelvis bone position (the $R^{\text{th}}$ joint of the body skeleton). We use $~\hat{}~$ to denote the predicted values. The $m^{\text{th}}$ human bounding box $\hat{\mathbf{B}}_{m}$ and root-relative 3D pose $\hat{\mathbf{P}}^{\mathit{rel}}_{m}$ are formulated as:
\begin{equation}
      \hat{\mathbf{B}}_{m} = (\hat{{u}}_{{m}}^{\mathit{top}}, \hat{{v}}_{{m}}^{\mathit{top}}, \hat{{w}}_{m}, \hat{{h}}_{m})^{\mathsf{T}},
\end{equation}
\begin{equation}
      \hat{\mathbf{P}}^{rel}_{m} = \{ (\hat{{u}}_{{m},j}, \hat{{v}}_{{m},j},\hat{{z}}^{\mathit{rel}}_{{m},j})^{\mathsf{T}} \}_{j=1}^J ,
\end{equation}
where $\hat{{u}}_{{m},j}$ and $\hat{{v}}_{{m},j}$ represent pixel coordinates of the estimated body joint with respect to the bounding box. $\hat{{z}}^{rel}_{{m},j}$ denotes the estimated depth of joint $j$ relative to the root-joint. $\hat{{u}}_{{m}}^{\mathit{top}}$, $\hat{{v}}_{{m}}^{\mathit{top}}$, $\hat{{w}}_{m}$, and $\hat{{h}}_{m}$ are the top left corner coordinates, the width, and the height of the predicted bounding box, respectively. With the intrinsic matrix $\mathbf{M}$, the final absolute 3D pose $\hat{\mathbf{P}}^{abs}_{m}$ can be obtained via back-projection, where each joint is calculated by:
\begin{equation}
   \left(
      \begin{array}{c}
         \hat{{x}}^{abs}_{{m},j} \\
         \hat{{y}}^{abs}_{{m},j} \\
         \hat{{z}}^{abs}_{{m},j}
      \end{array}
   \right) = (\hat{{z}}^{rel}_{{m},j} + \hat{{z}}^{abs}_{{m},R}) \mathbf{M}^{-1}  \left(
            \begin{array}{c}
               \hat{{u}}_{{m},j} + \hat{{u}}_{{m}}^{\mathit{top}} \\
               \hat{{v}}_{{m},j} + \hat{{v}}_{{m}}^{\mathit{top}}\\
               1
            \end{array}
         \right) .
   \label{eq:backproj}
\end{equation}

\begin{figure}[t]
   \begin{center}
      \includegraphics[width=.95\linewidth]{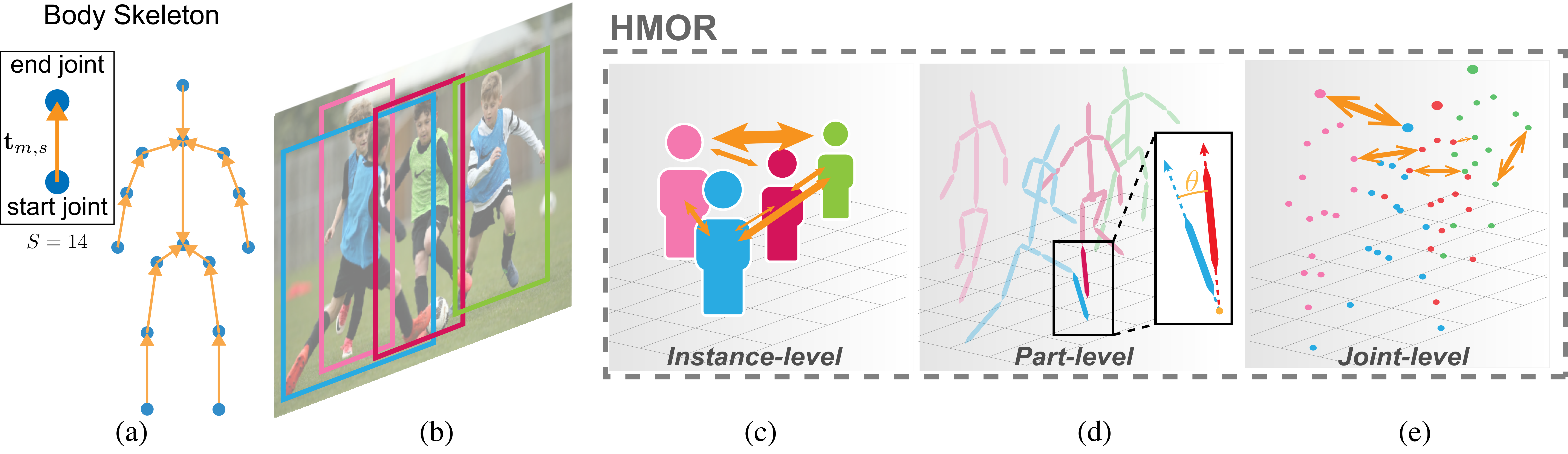}
   \end{center}
      \caption{{Illustration of the proposed HMOR.} (a) Definition of skeletal parts. (b) Monocular input image. (c-e) Hierarchical Multi-person Ordinal Relations. HMOR supervises the ordinal relations among multiple persons}
   \label{fig:hmor}
\end{figure}

\subsection{Hierarchical Multi-person Ordinal Relations}\label{sec:hmor}

Our initial goal is to leverage multi-person interaction relations to improve the performance of 3D pose estimation. Traditional top-down methods~\cite{lcr,lcrpp,moon} lack a global perspective because they estimate single human poses in each bounding box separately. Therefore, they are vulnerable to truncation, self-occlusions, and inter-person occlusions. Here, we develop a novel form of supervision named \textit{Hierarchical Multi-person Ordinal Relations (HMOR)} to model human relations explicitly. Basically, given an image of human activities, we divide the relationship into three levels: i) instance-level depth relations, ii) part-level angle relations, iii) joint-level depth relations. In each level, HMOR formulates pair-wise ordinal relations and punishes the wrong-order pairs. In the following, we detail our HMOR formulations that reflect interpretable relations of human activities.



\subsubsection{Instance-Level Depth Relations.}
In a given camera view, for two persons $(\mathbf{p}_1, \mathbf{p}_2)$, we denote the instance depth-relation function as $R_{\mathit{ins}}(\mathbf{p}_1, \mathbf{p}_2; \mathbf{n}_{\perp})$, taking the value:
\begin{itemize}
   \item[$\bullet$] $+1$, if $\mathbf{p}_1$ is closer than $\mathbf{p}_2$ in the $\mathbf{n}_{\perp}$ direction,
   \item[$\bullet$] $-1$, if $\mathbf{p}_2$ is closer than $\mathbf{p}_1$ in the $\mathbf{n}_{\perp}$ direction,
   \item[$\bullet$] $0$, if the depths of two person are equal,
\end{itemize}
where $\mathbf{n}_{\perp}$ is the camera normal vector.
We define the position of a person as the arithmetic mean of its body joints, i.e. $\mathbf{p}_m = \frac{1}{J} \sum_j^J \hat{\mathbf{k}}_{m,j}$. The ordinal error of a pair of instances is denoted as:
\begin{equation}
   {\mathit{err}}_{\mathit{ins}}(\hat{\mathbf{p}}_1, \hat{\mathbf{p}}_2) = \log(1 + \max(0, R_{\mathit{ins}}(\hat{\mathbf{p}}_1, \hat{\mathbf{p}}_2; \mathbf{n}_{\perp}) * [ (\hat{\mathbf{p}}_1 - \hat{\mathbf{p}}_2) \cdot \mathbf{n}_{\perp}])) .
   \label{eq:ins}
\end{equation}
This differentiable instance ranking expression will punish the wrong-order instance pairs and ignore the correct results. For example, if $\mathbf{p}_1$ is closer than $\mathbf{p}_2$, and the prediction relation is correct, i.e., $(\hat{\mathbf{p}}_1 - \hat{\mathbf{p}}_2) \cdot \mathbf{n}_{\perp} < 0$, the multiplication result will be smaller than $0$ and ignored by the maximum operation.

Supervising the instance-level depth relations is to help the network build a global understanding of the input scenario. Ablative study in \S\ref{sec:ablation} reveals that the accuracy of human-depth estimation benefits a lot from instance-level depth relations.

\subsubsection{Part-Level Angle Relations.}
As shown in Fig.~\ref{fig:hmor}(a), we divide the body skeleton into $S = 14$ parts according to the kinematically connected joints. Each part $\mathbf{t}$ is a vector defined by start-joint $\mathbf{k}_{\mathit{start}}$ and end-joint $\mathbf{k}_{\mathit{end}}$, i.e., $ \mathbf{t} = \mathbf{k}_{\mathit{end}} - \mathbf{k}_{\mathit{start}} $. Since body-parts are a set of 3D vectors with direction and length values, we can not directly compare their depths. Here, we utilize a unique attribute of body-part -- direction, and compare their angle relations. To simplify the ordinal relation of angles, we first project the body-part vector $\mathbf{t}_{m,s}$ onto the camera plane:
\begin{equation}
   \mathbf{t}^{\mathbf{n}_{\perp}}_{m,s} = \mathbf{t}_{m,s} - (\mathbf{t}_{m,s} \cdot \mathbf{n}_{\perp})\mathbf{n}_{\perp},
\end{equation}
where $m$ is the person index, and $s$ is the body-part index. In a given camera view, for a pair of body parts $(\mathbf{t}_{m_1, s_1}, \mathbf{t}_{m_2, s_2})$, we denote the angle-relation function as $R_{\mathit{arg}}(\mathbf{t}_{m_1, s_1}, \mathbf{t}_{m_2, s_2}; \mathbf{n}_{\perp})$, taking the value:
\begin{itemize}
   \item[$\bullet$] $+1$, if $\text{Arg}(\mathbf{t}_{m_1, s_1}^{\mathbf{n}_{\perp}}) < \text{Arg}(\mathbf{t}_{m_2, s_2}^{\mathbf{n}_{\perp}})$,
   \item[$\bullet$] $-1$, if $\text{Arg}(\mathbf{t}_{m_1, s_1}^{\mathbf{n}_{\perp}}) > \text{Arg}(\mathbf{t}_{m_2, s_2}^{\mathbf{n}_{\perp}})$,
   \item[$\bullet$] $0$, if $\text{Arg}(\mathbf{t}_{m_1, s_1}^{\mathbf{n}_{\perp}}) = \text{Arg} (\mathbf{t}_{m_2, s_2}^{\mathbf{n}_{\perp}})$,
\end{itemize}
where $\text{Arg}(\mathbf{t}^{\mathbf{n}_{\perp}})$ computes the principal value of the argument of the projection vector. The ordinal error of a pair of body-parts is:
\begin{equation}
   {\mathit{err}}_{\mathit{part}}(\hat{\mathbf{t}}_{m_1, s_1}, \hat{\mathbf{t}}_{m_2, s_2}) = [ R_{\mathit{arg}}(\hat{\mathbf{t}}_{m_1, s_1}, \hat{\mathbf{t}}_{m_2, s_2}; \mathbf{n}_{\perp}) * [(\hat{\mathbf{t}}_{m_1, s_1} \times \hat{\mathbf{t}}_{m_2, s_2}) \cdot \mathbf{n}_{\perp}] ]_{+} .
   \label{eq:part}
\end{equation}
With the cross-product operation $\times$, we supervise the direction of the angle between a pair of body-parts. If the angle between $\hat{\mathbf{t}}_{m_1, s_1}$ and $\hat{\mathbf{t}}_{m_2, s_2}$ is in the correct direction, the projection of the cross-product $(\hat{\mathbf{t}}_{m_1, s_1} \times \hat{\mathbf{t}}_{m_2, s_2}) \cdot \mathbf{n}_{\perp}$ will have an opposite sign of $R_{\mathit{arg}}(\cdot)$. Therefore, the negative multiplication results will be ignored by the $[\cdot]_+$ operation.

Another intuitive way is to express body-parts as particles and supervise their depth relations, using the average position of its two endpoints. To compare vector and particle representations, we conduct ablative experiments and find out vector is superior to particle representation. We suspect this is because the depth relations have been fully utilized in the other two levels, supervising depths of body-part is redundant. More experimental details are reported in \S\ref{sec:ablation}.

\subsubsection{Joint-Level Depth Relations.}
The definition of body joint depth-relation function $R_{\mathit{jt}}(\mathbf{k}_{m_1, s_1}, \mathbf{k}_{m_2, s_2}; \mathbf{n}_{\perp})$ is similar to $R_{\mathit{ins}}$:
\begin{itemize}
   \item[$\bullet$] $+1$, if $\mathbf{k}_{m_1, s_1}$ is closer than $\mathbf{k}_{m_2, s_2}$ in the $\mathbf{n}_{\perp}$ direction,
   \item[$\bullet$] $-1$, if $\mathbf{k}_{m_2, s_2}$ is closer than $\mathbf{k}_{m_1, s_1}$ in the $\mathbf{n}_{\perp}$ direction,
   \item[$\bullet$] $0$, if the depths of two joints are equal.
\end{itemize}
The ordinal error of a pair of joints is denoted as:
\begin{equation}
   {\mathit{err}}_{\mathit{jt}}(\hat{\mathbf{k}}_{m_1, s_1}, \hat{\mathbf{k}}_{m_2, s_2}) = \log(1 + [R_{\mathit{jt}}(\hat{\mathbf{k}}_{m_1, s_1}, \hat{\mathbf{k}}_{m_2, s_2}; \mathbf{n}_{\perp}]_+ * [ (\hat{\mathbf{k}}_{m_1, s_1} - \hat{\mathbf{k}}_{m_2, s_2}) \cdot \mathbf{n}_{\perp}])) .
   \label{eq:joint}
\end{equation}
Denoting the set of estimated \textit{persons}, \textit{body-parts}, and \textit{joints} pairs as $\mathcal{I}_{\mathit{ins}}$, $\mathcal{I}_{\mathit{part}}$, and $\mathcal{I}_{\mathit{jt}}$, respectively, the HMOR loss is computed as follows:
\begin{equation}
   \mathcal{L}_{\mathit{HMOR}}
   =
   \frac{1}{|\mathcal{I}_{\mathit{ins}}|} \sum_{\hat{\mathbf{p}}_1, \hat{\mathbf{p}}_2} {\mathit{err}}_{\mathit{ins}}
   +
   \frac{1}{|\mathcal{I}_{\mathit{part}}|} \sum_{\hat{\mathbf{t}}_1, \hat{\mathbf{t}}_2} \mathit{err}_{\mathit{part}}
   +
   \frac{1}{|\mathcal{I}_{\textit{jt}}|} \sum_{\hat{\mathbf{k}}_1, \hat{\mathbf{k}}_2} \mathit{err}_{\mathit{jt}}
\end{equation}

\noindent\textbf{Augmented Training Scheme.}~
As mentioned before, HMOR computes the ordinal relations with respect to a vector $\mathbf{n}_{\perp}$. Initially, this vector is set as the camera normal vector. However, we notice that annotations from 3D human pose datasets (Human3.6M, MuPoTS-3D, and CMU Panoptic) are mostly captured in an laboratory environment, limited to the fixed viewing angle. To alleviate camera restrictions, we sample virtual views to improve the generalization ability.

In the training phase, we generate a virtual view vector $\mathbf{n}_{v}$ by rotating the camera normal vector $\mathbf{n}_{\perp}$ randomly. We adapt the uniform sphere sampling strategy from Marsaglia et al.~\cite{sphere}:
\begin{equation}
   \mathbf{n}_{v} = (\sqrt{1 - u^2} \cos{\theta}, \sqrt{1 - u^2} \sin{\theta}, u)^{\mathsf{T}},
\end{equation}
where $\theta \sim U[0, 2\pi)$ and $u \sim U[0, 1]$. In this way, HMOR can calculate the ordinal relations with respect to an arbitrary viewing angle. The effectiveness of the sampled view is validated in \S\ref{sec:ablation}.

Additionally, a mixed datasets training strategy is utilized for a fair comparison with previous methods in experiments. HMOR is compatible with 2D pose datasets and single-person 3D pose datasets. Given an image only with 2D pose annotations, we can define the part-level angle relations, since the 2D pose skeletons are the projections of body-parts with respect to $\mathbf{n}_{\perp}$. As for single-person cases, HMOR only supervises the \textit{joint} and \textit{body-part} relations of an individual person and ignore instance-level relations.

\subsection{Integrated End-to-end Model}\label{sec:end2end}

\begin{figure}[t]
   \begin{center}
   \includegraphics[width=.95\linewidth]{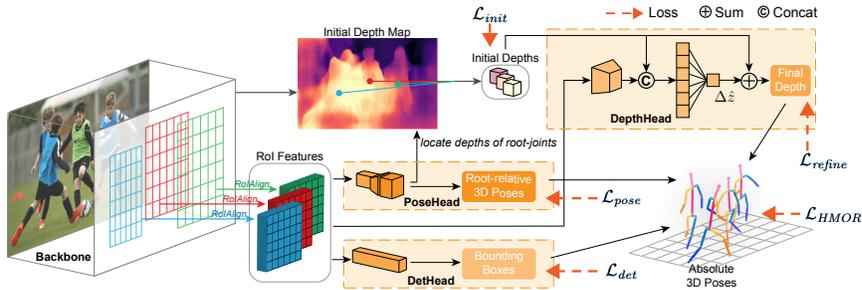}
   \end{center}
      \caption{{Architecture of the integrated model.} The ResNet-50 based backbone network extract RoI features and initial depth map. PoseHead and DetHead perform root-relative 3D pose estimation and human detection, respectively. DepthHead retrieves initial depths from the depth map and predicts refined human depths by correction residual $\Delta\hat{z}$. This architecture allows the 2D-3D mixed training strategy
      }
   \label{fig:pipeline}
\end{figure}

In our approach, an integrated end-to-end top-down model is designed to aggregate HMOR into the end-to-end training process. Although the disjoint model~\cite{moon} can use different strong networks for different tasks (e.g., human detection, pose estimation, depth estimation),  an integrated model has three advantages over the disjoint learning model: 1) Fewer model parameters. 2) Only an integrated model can leverage the multi-person relations since the disjoint learning methods train their model with single person annotations separately. 3) The multi-task training strategy of the integrated model can benefit each task. In our experiments, the integrated model is found to have much better performance than the disjoint learning methods.

The overall architecture of our model is summarized in Fig.~\ref{fig:pipeline}. Our model consists of two stages. In the first stage, the backbone network extracts RoIs and the initial depth map. PoseHead and DetHead estimate root-relative 3D poses and human bounding boxes from RoIs, respectively. In the second stage, we retrieve the initial depths of root-joints from the depth map. The DepthHead takes the RoI features and initial depth as input and outputs the correction residual $\Delta\hat{z}$. The residual is added to the initial depth to obtain refined human depths. Aggregating the outputs from DetHead, PoseHead, and DepthHead, the absolute 3D poses $\hat{\mathbf{P}}^{abs}$ are estimated via back-projection as Eq.~\ref{eq:backproj}.

\subsubsection{Human Detection.}
The architecture for human detection and the loss function $L_{\mathit{det}}$ follow the design in Mask R-CNN~\cite{maskrcnn}. Region Proposal Network (RPN) proposes candidate human bounding boxes, and the DetHead predicts class labels and bounding-box offsets. RoiAlign is used to extract feature maps from each RoI.

\subsubsection{Root-Relative 3D Pose Estimation.}
PoseHead is proposed to estimate the root-relative 3D pose $\hat{\mathbf{P}}^{rel}$ from an input RoI feature. We use 3D heatmaps as the representations of 3D poses. The soft-argmax operation~\cite{integral} is adopted to extract $\hat{\mathbf{P}}^{rel}$ from the 3D heatmap. $\ell_1$ regression loss is applied to root-relative coordinates $\hat{\mathbf{P}}^{rel}_m$:
\begin{equation}
   \mathcal{L}_{\mathit{pose}} = \frac{1}{N} \frac{1}{J} \sum_m^N \| \hat{\mathbf{P}}_m^{\mathit{rel}} - \mathbf{P}_m^{\mathit{rel}} \|_1 .
   \label{eq:loss_reg}
\end{equation}
RoIAlign extracts $14 \times 14$ RoI features, which are fed into the PoseHead subsequently. We adopt a simple network as PoseHead, including three residual blocks for feature extraction, a transposed convolution~\cite{transposed} for upsampling, a batch normalization layers\cite{batchnorm}, a ReLU activation function, and a $1\times1$ convolution. The size of an output heatmap is $28 \times 28 \times 28$.

\subsubsection{Human Depth Estimation.}
Direct human-depth regression from an input RoI is challenging. Part of the challenges comes from the variety of camera parameters and human body shapes. Furthermore, the inputs of DepthHead are fixed-size RoI features, which erase the information of projected body shapes and sizes. Inspired by the idea of iterative error feedback (IEF) from previous works~\cite{ief1,ief2,kanazawa2018end}, we design a coarse-to-fine estimation approach to enhance the accuracy of human-depth regression. The model will first predict an initial depth of root-joint $\hat{z}^{\mathit{init}}$. Then the DepthHead takes the RoI features and the initial depth $\hat{z}^{\mathit{init}}$ as an input and outputs the residual $\Delta{z}$. Ideally, the refined depth is updated by adding this residual to the initial estimate $\hat{z}^{\mathit{refine}} = \hat{z}^{\mathit{init}} + \Delta{z}$.

\paragraph{Depth Initialize.} To estimate the initial depths of root-joints, we directly regress an initial depth map. During training, we first normalize the absolute depth value by focal lengths and then calculate the loss $\mathcal{L}_{\mathit{init}}$ between the ground truth and the initial depth map in the area around the root-joint's 2D pixel location:
\begin{equation}
   {z}_{R}^{\mathit{norm}} = {{z}_R^{abs}} / {\sqrt{f_x \cdot f_y}}.
\end{equation}
\begin{equation}
   \mathcal{L}_{\mathit{init}} = \frac{1}{N} \sum_m^N \| {{{z}}_{m,R}^{\mathit{norm}}} - \hat{z}_{m,R}^{\mathit{init}} \|_1 ,
   \label{eq:loss_map}
\end{equation}

\paragraph{Depth Refinement.}
In the refinement step, we retrieve the initial-depth values of root-joints from the depth map according to their 2D pixel locations. Because the input features are resized by RoIAlign, we first need to transfer the original depth to the equivalent depth of the resized person. According to the pinhole camera model:
\begin{equation}
   z_{R}^{\mathit{eq}, \mathit{norm}} = z_{R}^{\mathit{norm}} \cdot \sqrt{\frac{A_{\mathit{Box}}}{A_{\mathit{RoI}}}} ,
\end{equation}
\begin{equation}
   \hat{z}_{R}^{\mathit{eq}, \mathit{init}} = \hat{z}_{R}^{\mathit{init}} \cdot \sqrt{\frac{A_{\mathit{Box}}}{A_{\mathit{RoI}}}} ,
\end{equation}
where $A_{\mathit{Box}}$ denotes the area of the bounding box, and $A_{\mathit{RoI}}$ denotes the area of RoI. DepthHead extracts 1D features from RoIs. Then the equivalent initial-depth values $\hat{{z}}_{m,R}^{\mathit{eq}, \mathit{init}}$ are concatenated with the extracted features and fed into an \textit{fc} layer to predict the residual $\Delta\hat{z}$.
The loss function of the refinement step $\mathcal{L}_{\mathit{refine}}$ is defined as:
\begin{equation}
   \mathcal{L}_{\mathit{refine}} = \frac{1}{N} \sum_m^N \| {z}_{m,R}^{\mathit{eq},\textit{norm}} - \hat{{z}}_{m,R}^{\mathit{eq},\textit{init}} - \Delta\hat{z}_m  \|_1 .
   \label{eq:loss_root}
\end{equation}
In the testing phase, we can recover the absolute depth of root-joint $\hat{\mathbf{z}}^{abs}_{m,R}$ as:
\begin{equation}
   \hat{\mathbf{z}}^{abs}_{m,R} = (\Delta\hat{z}_m + {\hat{\mathbf{z}}_{m,R}^{eq,\mathit{init}}}) \cdot \sqrt{\frac{{f_x \cdot f_y \cdot A_{\mathit{RoI}}}}{A_{\mathit{Box}}}} .
   \label{eq:norm_abs}
\end{equation}

The DepthHead uses three residual blocks (following ResNet~\cite{resnet}) and an average pooling layer to extract 1D features. The FC layer contains $512$ neurons.

The end-to-end training loss is formulated as:
\begin{equation}
   \mathcal{L} = \mathcal{L}_{\mathit{det}}+\mathcal{L}_{\mathit{pose}} + \mathcal{L}_{\mathit{init}} + \mathcal{L}_{\mathit{refine}}  + \mathcal{L}_{\mathit{HMOR}} .
\end{equation}

\section{Experiment}
In this section, we first introduce the datasets employed for quantitative evaluation and elaborate implementation details. Then we report our results and compare the proposed method with state-of-the-art methods. Finally, ablation experiments are conducted to evaluate our contributions and show how each choice contributes to our state-of-the-art performance.

\subsection{Datasets}

\noindent\textbf{MuCo-3DHP and MuPoTS-3D:}\quad
{MuCo-3DHP} is a multi-person composited 3D human pose training dataset. {MuPoTS-3D} is the real-world scenes test set. Following~\cite{singleshot,moon}, 400K composited frames are utilized for training.

\noindent\textbf{CMU Panoptic.}\quad
CMU Panoptic~\cite{panoptic} is a multi-person 3D pose dataset captured in an indoor dome with multiple cameras. Here we follow the evaluation protocol of \cite{zanfir2018monocular,nips2018}.

\noindent\textbf{3DPW.}\quad
3D Poses in the Wild (3DPW)~\cite{3dpw} is a recent challenging dataset, captured mostly in outdoor conditions. It contains 60 video sequences (24 train, 24 test, and 12 validation).

\noindent\textbf{Human3.6M.}\quad
Human3.6M~\cite{h36m} is an indoor benchmark for single-person 3D pose estimation. A total of 11 professional actors (6 male, 5 female) perform 15 activities in a laboratory environment.


\subsection{Implementation Details}
Our method is implemented in PyTorch. We adopt a ResNet-50~\cite{resnet} based FPN~\cite{fpn} as our model backbone. The backbone is initialized with the ImageNet~\cite{imagenet} pre-trained model. The settings of each network head are reported in \S\ref{sec:end2end}. We resize the image to $1333 \times 800$ and feed into the network. SGD is used for optimization, with a mini-batch size of $32$. All tasks are trained simultaneously. We adopt the linear learning rate warm-up policy. The learning rate is set to $0.2/3$ at first and gradually increases to $0.2$ after 2.5k iterations. We reduce the learning rate by a factor of $10$ at the 10th and 20th epochs. In each experiment, our model is trained for 30 epochs with 16 NVIDIA 1080 Ti GPUs. We perform data augmentations including horizontal flip and multi-scale training. Additional COCO~\cite{mscoco} 2D pose estimation data are used in the training phase. For evaluation, we report the flip-test results. All reported numbers have been obtained with a single model without ensembling.

\subsection{Compare with Prior Art}

\begin{table}[t]
    \begin{center}
        \caption{Quantitative comparisons with state-of-the-art methods on the MuPoTS-3D dataset. ``-'' shows the results that are not available}
        \label{table:mupots}
    
        \begin{tabular}{l|ccc}
		\toprule

        Method & ~AUC$_{rel}$$\uparrow$~ & ~3DPCK$_{rel}$$\uparrow$~ & ~3DPCK$_{abs}$$\uparrow$~ \\
		\midrule

        \cellcolor{Gray}LCRNet~\cite{lcr}~ & \cellcolor{Gray}- & \cellcolor{Gray}53.8 & \cellcolor{Gray}- \\
        Single Shot~\cite{singleshot}~ &  - &  66.0 &  - \\
        \cellcolor{Gray}LCRNet++~\cite{lcrpp}~ & \cellcolor{Gray}- & \cellcolor{Gray}70.6 & \cellcolor{Gray}- \\
        Xnect~\cite{xnect}~ &  - &  70.4 &  - \\
        \cellcolor{Gray}Moon et al.~\cite{moon}~ & \cellcolor{Gray}39.8 & \cellcolor{Gray}81.8 & \cellcolor{Gray}31.5 \\
		\midrule

        {Ours} &  \textbf{43.5} &  \textbf{82.0} &  \textbf{43.8} \\
		\bottomrule
        \end{tabular}
    \end{center}
\end{table}

\begin{table}[t]
\begin{center}
    \caption{Quantitative comparisons of MPJPE on the CMU Panoptic dataset}
    \label{table:panoptic}
    \begin{tabular}{l|ccccc}
    \toprule
    Method & ~Haggling~ & ~Mafia~ & ~Ultimatum~ & ~Pizza~ & ~Mean$\downarrow$~ \\
    \midrule
    \cellcolor{Gray}Popa~\cite{popa2017deep}~ & \cellcolor{Gray}217.9 & \cellcolor{Gray}187.3 & \cellcolor{Gray}193.6 & \cellcolor{Gray}221.3 & \cellcolor{Gray}203.4 \\
    Zanfir~\cite{zanfir2018monocular}~ & 140.0 & 165.9 & 150.7 & 156.0 & 153.4 \\
    \cellcolor{Gray}Zanfir~\cite{nips2018}~ & \cellcolor{Gray}72.4 & \cellcolor{Gray}78.8 & \cellcolor{Gray}66.8 & \cellcolor{Gray}94.3 & \cellcolor{Gray}72.1\\
    \midrule
    {Ours} & \textbf{50.9} & \textbf{50.5} & \textbf{50.7} & \textbf{68.2} & \textbf{51.6} \\
    \bottomrule
    \end{tabular}
\end{center}
\end{table}

\begin{table}[t]
\begin{center}   
    \caption{Quantitative comparisons on the Human3.6M dataset}
    \label{table:h36m}
    \setlength\tabcolsep{1.0pt}
    \begin{tabular}{l|cccccccccc}
    \toprule
    & \multicolumn{7}{l}{{\textit{Single-Person}}} & \multicolumn{3}{|l}{{\textit{Multi-Person}}} \\\hline
    Method & \tabincell{c}{Moreno\\\cite{twostage3}} & \tabincell{c}{Zhou\\\cite{zhou2018monocap}} & \tabincell{c}{Martinez\\\cite{martinez2017simple}} & \tabincell{c}{Sun\\\cite{compositional}} & \tabincell{c}{Fang\\\cite{fang2018}} & \tabincell{c}{Sun\\\cite{integral}} & \tabincell{c}{Zhou\\\cite{zhou2019hemlets}} &
    \multicolumn{1}{|c}{\tabincell{c}{Rogez\\\cite{lcrpp}}} & \tabincell{c}{Moon\\\cite{moon}} & \tabincell{c}{\textbf{Ours}} \\
    \hline
    {PA MPJPE}$\downarrow$~ & 76.5 & 55.3 & 47.7 & 48.3 & 45.7
    & {40.6} & \textbf{27.9}
    & \multicolumn{1}{|c}{42.7} & 35.2 & \textbf{30.5} \\

    \midrule

    & \multicolumn{5}{l}{{\textit{Single-Person}}} & \multicolumn{5}{|l}{{\textit{Multi-Person}}} \\\hline
    Method & \tabincell{c}{Martinez\\\cite{martinez2017simple}} & \tabincell{c}{Fang\\\cite{fang2018}} & \tabincell{c}{Sun\\\cite{compositional}} & \tabincell{c}{Sun\\\cite{integral}} & \tabincell{c}{Zhou\\\cite{zhou2019hemlets}} &
    \multicolumn{1}{|c}{\tabincell{c}{Rogez\\\cite{lcr}}} & \tabincell{c}{Metha\\\cite{singleshot}} & \tabincell{c}{Rogez\\\cite{lcrpp}} & \tabincell{c}{Moon\\\cite{moon}} & \tabincell{c}{\textbf{Ours}} \\
    \hline
    {MPJPE}$\downarrow$~ & 62.9 & 60.4 & 59.1 & 49.6 & \textbf{39.9} &
    \multicolumn{1}{|c}{87.7} & 69.9 & 63.5 & 54.4 & \textbf{48.6} \\
    \bottomrule
    \end{tabular}
\end{center}
\end{table}

\noindent\textbf{MuPoTS-3D.}\quad
We compare our method against state-of-the-art methods under three protocols. {PCK}$_{\mathit{abs}}$ is used to evaluate absolute camera-centered coordinates of 3D poses. Additionally, {PCK}$_{\mathit{rel}}$ and {AUC}$_{rel}$ are used to evaluate root-relative 3D poses after root alignment. Quantitative results are reported in Table~\ref{table:mupots}. Without bells and whistles, our method surpasses state-of-the-art methods by {12.3} {PCK}$_{\mathit{abs}}$ ($\mathbf{39.0} \%$ relative improvement). Our method demonstrates a clear advantage for handling multi-person 3D poses.

As for root-relative results, our method achieves {82.0} {PCK}$_{\mathit{rel}}$ and {43.5} {AUC}$_{rel}$. Note that the PCK result relies on the threshold value. AUC can reflect a more reliable result since it computes the area under the PCK curve from various thresholds. Our method outperforms the previous methods by 3.7 {AUC}$_{rel}$.

~\\
\noindent\textbf{CMU Panoptic.}\quad
Following previous works~\cite{zanfir2018monocular,nips2018}, we evaluate our method under MPJPE after root alignment. Table~\ref{table:panoptic} provides experimental results. In this dataset, the activities take place in a small room. Thus, the scenarios are severely affected by the occlusion problem. Our method effectively reduces the interference of occlusion and outperforms state-of-the-art methods by {20.5} mm {MPJPE} ($\mathbf{28.4} \%$ relative improvement).

~\\
\noindent\textbf{Human3.6M.}\quad
We conduct experiments on Human3.6M dataset to evaluate the performance of root-relative 3D pose estimation. Two experimental protocols are widely used. \textit{Protocol 1} uses \textit{PA MPJPE} and \textit{Protocol 2} uses \textit{MPJPE} as evaluation metrics. As most of the previous methods use the ground-truth bounding box, our method does not require any ground-truth information at inference time. Quantitative results are reported in Table~\ref{table:h36m}. Our method achieves comparable performance with single-person methods and outperforms previous multi-person 3D pose estimation methods.


\begin{table}[t]
    \renewcommand{\arraystretch}{1}
    \begin{center}
    \caption{Ablative study on the effects of HMOR}
    \label{table:ablation}
    \begin{tabular}{cl|ccc}
    \toprule
    \multicolumn{2}{c|}{\multirow{2}{*}{Settings}} & \multicolumn{3}{c}{3DPW} \\\cmidrule(r){3-5}
    & & ~MPJPE$\downarrow$~ & ~PA- $\downarrow$~ & ~ABS- $\downarrow$~\\
    \midrule
    \multirow{9}{*}{(a)}~ & \cellcolor{Gray}baseline & \cellcolor{Gray}95.7 & \cellcolor{Gray}63.6 & \cellcolor{Gray}169.3 \\
    & + $\mathcal{L}_{\mathit{abs}}$ & 94.6 & 61.1 & 158.2 \\
    & \cellcolor{Gray}+ \textit{jt} & \cellcolor{Gray}89.9 & \cellcolor{Gray}59.7 & \cellcolor{Gray}132.8 \\
    & + \textit{part} & 90.2 & 60.3 & 143.2\\
    & \cellcolor{Gray}+ \textit{instance} & \cellcolor{Gray}93.3 & \cellcolor{Gray}61.2 & \cellcolor{Gray}128.3 \\
    & + \textit{jt} + \textit{part} & 89.1 & 58.3 & 125.9 \\
    & \cellcolor{Gray}+ \textit{jt} + \textit{instance} & \cellcolor{Gray}89.2 & \cellcolor{Gray}58.5 & \cellcolor{Gray}122.3 \\
    & + \textit{part} + \textit{instance} & 89.5 & 59.5 & 123.6 \\
    & \cellcolor{Gray}+ \textit{jt} + \textit{part} + \textit{instance} & \cellcolor{Gray}88.3 & \cellcolor{Gray}57.8 & \cellcolor{Gray}119.6\\
    \midrule
    (b)~ & + \textit{jt} + \textit{particle-part} + \textit{instance} & 89.0 & 58.2 & 119.5 \\
    \midrule
    (c)~ & \cellcolor{Gray}+ \textit{jt} + \textit{part} + \textit{instance} + \textit{sample views} (Final)~ & \cellcolor{Gray}\textbf{87.7} & \cellcolor{Gray}\textbf{57.4} & \cellcolor{Gray}\textbf{118.5} \\
    \midrule
    (d)~ & w/o refine depth & 88.4 & 58.1 & 133.6 \\
    \bottomrule
    \end{tabular}
    \end{center}
\end{table}

\subsection{Ablation Study}
\label{sec:ablation}
In this study, we evaluate the effectiveness of the proposed HMOR and integrated model. We evaluate on 3DPW dataset that contains in-the-wild complex scenes to demonstrate the strength of our model. We further propose ABS-MPJPE to evaluate the absolute 3D pose estimation results without root alignment.

\noindent\textbf{Effect of Hierarchical Multi-person Ordinal Relations.}\quad
In this experiment, we study the effectiveness of using HMOR supervision. We first implement a vanilla baseline without HMOR supervision. Moreover, we implement another baseline by directly supervising the predicted absolute 3D poses with an $\ell_1$ loss $\mathcal{L}_{\mathit{abs}}$. Intuitively, since the human poses are evaluated in the camera coordinate system, the local optimum for $\mathcal{L}_{\mathit{abs}}$ is consistent with the evaluation metrics.

The experimental results are shown in Table~\ref{table:ablation}(a). The model trained with $\mathcal{L}_{\mathit{abs}}$ supervision has better performance than the vanilla baseline, but is still inferior to HMOR supervision. HMOR supervision brings 7.4 mm MPJPE improvement. By removing three types of relations separately, we can observe that \textit{instance} relation affects the absolute pose accuracy (ABS-MPJPE) most, while \textit{part} and \textit{joint} relations mainly affect the root-relative pose accuracy.

~\\
\noindent\textbf{Variants of HMOR.}\quad
In this experiment, we examine a variant of HMOR. When handling the part relations, we represent a body part as a particle rather than a vector. The position of a body part is defined as the average of its two endpoints. Similar to \textit{joint} and \textit{instance}, we supervise the depth relations of the particle body-parts. The experimental results are shown in Table~\ref{table:ablation}(b). The particle representation produces inferior performance than the vector representation.

~\\
\noindent\textbf{Effect of Sampled Views.}\quad
Table~\ref{table:ablation}(c) reports the result of training with sampled views. Compare with the results in Table~\ref{table:ablation}(a) that only use the original camera normal vector $\mathbf{n}_{\perp}$, sampled views provide 0.6 mm MPJPE improvement.

~\\
\noindent\textbf{Effect of Coarse-to-Fine Depth Surpervision.}\quad
In this experiment, we study the effectiveness of the coarse-to-fine design for human depth estimation. We remove the refinement step and output the initial value directly. The experimental results are shown in Table~\ref{table:ablation}(d). We observe that the coarse-to-fine design is necessary to produce accurate human depth.

\begin{figure}[t]
    \begin{center}
    \includegraphics[width=.95\linewidth]{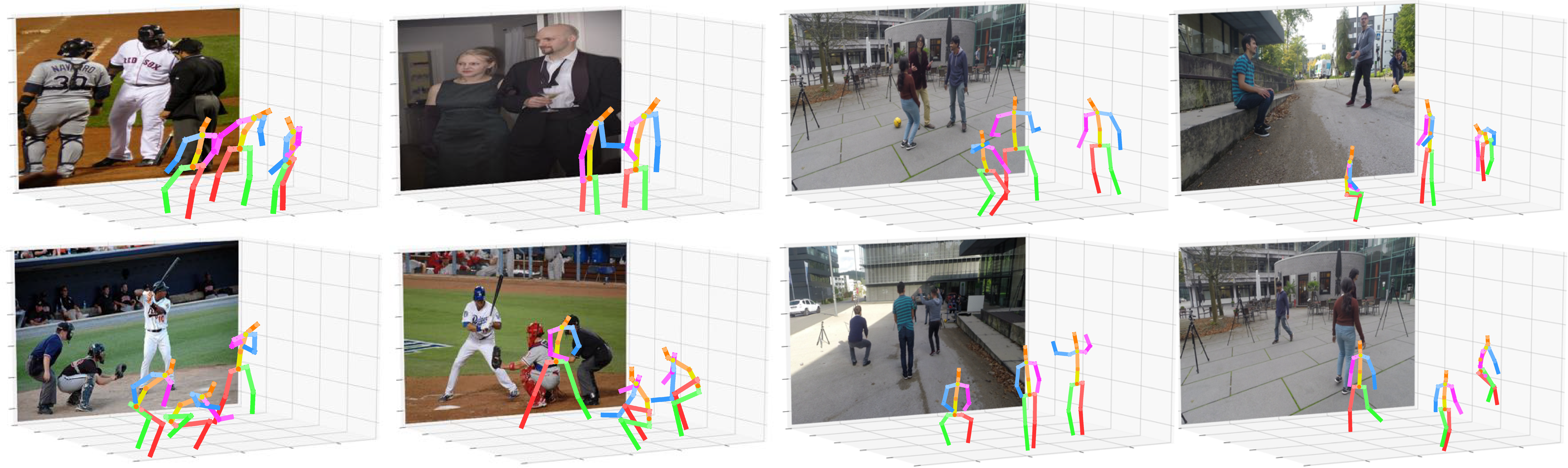}
    \end{center}
    \caption{Qualitative results of our proposed method on COCO validation set (left) and MuPoTS-3D test set (right)}
    \label{fig:qualitative results}
\end{figure}

\begin{table}[t]
    \begin{center}
    \caption{Ablative study on computation complexity and model parameters}
    \label{table:complexity}
    \begin{tabular}{l|ccccc}
    \toprule
    Method & ~\#Params$\downarrow$~ & ~GFLOPs$\downarrow$~ & ~\tabincell{c}{AUC$_{\mathit{rel}}$$\uparrow$}~ & ~\tabincell{c}{PCK$_{\mathit{rel}}$$\uparrow$}~ & ~\tabincell{c}{PCK$_{\mathit{abs}}$$\uparrow$}~ \\
    \midrule
    \cellcolor{Gray}Moon~\cite{moon} & \cellcolor{Gray}167.7M & \cellcolor{Gray}547.8 & \cellcolor{Gray}39.8 & \cellcolor{Gray}81.8 & \cellcolor{Gray}31.5 \\
    Ours & \textbf{45.0M} & \textbf{320.2} & \textbf{43.5} & \textbf{82.0} & \textbf{43.8} \\
    \bottomrule
    \end{tabular}
    \end{center}
\end{table}

\noindent\textbf{Computation Complexity.}\quad
The experimental results of computation complexity and model parameters are listed in Table~\ref{table:complexity}. We compare our method with Moon et al.~\cite{moon}, which is the only open-source multi-person 3D pose estimation method. Our approach obtains superior results to the state-of-the-art 3D pose estimation method (both absolute pose and root-relative pose), with significantly lower computation complexity and fewer model parameters.

\section{Conclusion}

In this paper, we proposed a novel form of supervision - HMOR, to learn multi-person 3D poses from a monocular RGB image. HMOR supervises the multi-person ordinal relations in a hierarchical manner, which captures fine-grained semantics and maintains global consistency at the same time. To end-to-end learn the ordinal relations, we further proposed an integrated model with a coarse-to-fine depth-estimation architecture. We demonstrate the effectiveness of our proposed method on standard benchmarks. The proposed method surpasses state-of-the-art multi-person 3D pose estimation methods, with lower computation complexity and fewer model parameters. We believe the idea of leveraging multi-person relations can be further explored to improve 3D pose estimation, e.g., exploit the relations via network design.

\subsubsection{Acknowledgements.}
This work is supported in part by the National Key R\&D Program of China, No. 2017YFA0700800, National Natural Science Foundation of China under Grants 61772332Shanghai Qi Zhi Institute.

\clearpage
%
%
\bibliographystyle{splncs04}
\bibliography{egbib}
\end{document}